\documentclass[runningheads]{llncs}

\usepackage[T1]{fontenc}
\usepackage{graphicx}
\graphicspath{{Data/imgs/}}
\usepackage{booktabs}
\usepackage{multirow}
\usepackage{amsmath,amssymb}
\usepackage{mathtools}
\usepackage{microtype}
\usepackage{xcolor}
\usepackage{url}
\usepackage{cite}
\usepackage{hyperref}
\usepackage{enumitem}
\usepackage{array}
\usepackage{tabularx}
\usepackage{siunitx}

\hypersetup{
    colorlinks=true,
    linkcolor=blue,
    citecolor=blue,
    urlcolor=blue,
    pdfauthor={Athanasios Angelakis},
    pdftitle={hZACH-ViT: Curved Latent Geometry for Compact Vision Transformers in Low-Data Medical Imaging},
    pdfsubject={Medical Imaging, Vision Transformers, Riemannian Geometry},
    pdfkeywords={Vision Transformers, Hyperbolic learning, Riemannian manifolds, Medical imaging}
}

\title{hZACH-ViT: Curved Latent Geometry for Compact Vision Transformers in Low-Data Medical Imaging}

\titlerunning{hZACH-ViT}

\author{Athanasios Angelakis\inst{1,2}}
\authorrunning{A. Angelakis}
\institute{BioML Lab, Research Institute CODE, UniBw, Munich, Germany\\
\email{athanasios.angelakis@unibw.de}
\and
Department of Epidemiology and Data Science, Amsterdam UMC, Amsterdam, Netherlands}

\begin{document}
\maketitle

\begin{abstract}
Compact Vision Transformers are attractive for medical imaging in low-data and resource-constrained settings, but most existing variants assume that Euclidean latent geometry is sufficient for organizing image representations. We introduce hZACH-ViT, a family of curved-geometry extensions of ZACH-ViT, a compact zero-token Vision Transformer that removes positional embeddings and the class token and relies on global average pooling over patch representations. To isolate the role of geometry, we preserve the verified ZACH-ViT backbone and modify only the final representation space and prototype-based classifier head, enabling a controlled comparison between Euclidean, hyperbolic, and spherical latent geometries.

We evaluate Poincar\'e, Klein, and spherical hZACH-ViT heads on seven MedMNIST datasets under an identical few-shot protocol with 50 samples per class and five random seeds. The completed benchmark contains 770 training runs spanning seven datasets, three non-Euclidean geometries, seven curvature magnitudes, and a Euclidean baseline. Across all seven datasets, the best non-Euclidean hZACH-ViT configuration improves over Euclidean ZACH-ViT, with an average gain of $+0.021$ in the dataset-specific primary metric and the largest improvement on OCTMNIST ($+0.055$ MacroF1). Fixed low-curvature configurations retain positive gains on the majority of datasets, and low curvature values ($c=0.1$ or $0.2$) account for six of the seven dataset-level winners.

Rather than identifying a universally optimal manifold, our results establish geometry and curvature as dataset-dependent model-selection variables, with fixed low-curvature analyses confirming that gains persist beyond exhaustive per-dataset tuning.
\end{abstract}

\section{Introduction}

Vision Transformers (ViTs) have become central to modern computer vision, yet their success often depends on architectural assumptions inherited from large-scale natural image modeling, including positional embeddings, token-based aggregation, and Euclidean latent representations~\cite{dosovitskiy2021image}. In medical imaging, these assumptions may not always be optimal. Several medical image modalities are characterized by limited annotated data, acquisition variability, weak or inconsistent global spatial layout, and deployment constraints that favor compact architectures over large pretrained models.

ZACH-ViT was introduced as a compact permutation-invariant Vision Transformer for medical imaging~\cite{angelakis2026zachvit}. It removes both positional embeddings and the dedicated class token, using global average pooling over patch representations instead. This design makes the model explicitly zero-token with respect to aggregation and positional priors, while still processing patch tokens normally. Prior evaluation across seven MedMNIST datasets~\cite{yang2023medmnist} showed that ZACH-ViT is particularly competitive in weak spatial-structure regimes, such as blood microscopy and histopathology, and that the usefulness of positional priors is regime dependent~\cite{angelakis2026zachvit}.

The present work asks a complementary question: once spatial priors and class-token aggregation are removed, is Euclidean latent geometry still the most appropriate choice? In many representation-learning problems, non-Euclidean geometries provide useful inductive biases. Hyperbolic geometry is well suited for hierarchical or tree-like structures~\cite{nickel2017poincare,ganea2018hyperbolic}, while spherical geometry may better represent clustered or angular class organization~\cite{liu2017sphereface,davidson2018hyperspherical}. Medical image classes may contain hierarchical, compositional, or manifold-like relationships that are not optimally captured in a flat Euclidean latent space.

We therefore introduce hZACH-ViT, a geometry-aware extension of ZACH-ViT designed to test whether curved latent spaces can systematically improve compact medical image classification. The core principle is to keep the verified ZACH-ViT backbone unchanged and modify only the final latent geometry and classifier head. This design isolates the role of representation geometry from other architectural factors. We study three non-Euclidean variants: Poincar\'e hyperbolic hZACH-ViT, Klein hyperbolic hZACH-ViT, and spherical hZACH-ViT. Curvature is treated as an explicit experimental variable, while all backbone, optimization, sampling, and evaluation settings are fixed. We complete a 770-run benchmark across seven MedMNIST datasets, five random seeds, three non-Euclidean geometries, seven curvature magnitudes, and a Euclidean baseline.\footnote{Code, execution notebooks, and aggregated result summaries will be released at \url{https://github.com/Bluesman79/hZACH-ViT} upon publication.}

The contributions of this work are:
\begin{itemize}[leftmargin=*]
    \item We introduce hZACH-ViT, a family of curved-latent-space extensions of the compact zero-token ZACH-ViT architecture.
    \item We design a controlled comparison in which the verified ZACH-ViT backbone is kept fixed and only the final geometry-aware prototype head is changed, isolating the effect of latent geometry from confounding architectural changes.
    \item We complete a 770-run benchmark across seven MedMNIST datasets, five random seeds, three non-Euclidean geometries, seven curvature magnitudes, and a Euclidean baseline.
    \item We show that the best hZACH-ViT configuration improves over Euclidean ZACH-ViT on all seven datasets, while also demonstrating that the preferred geometry and curvature are dataset dependent rather than universal.
    \item We provide manuscript-ready 300-DPI figures and a transparent aggregation protocol, making the geometry--curvature behavior of hZACH-ViT easy to inspect and extend.
\end{itemize}

\section{Related Work}

\subsection{Compact Vision Transformers for Medical Imaging}

Vision Transformers were originally developed for large-scale image recognition, where abundant data and strong spatial structure make positional encoding and token-based aggregation effective~\cite{dosovitskiy2021image}. In medical imaging, however, data scarcity and heterogeneous acquisition protocols often motivate smaller models and stronger inductive-bias control. Compact transformers and hybrid architectures have therefore become attractive for low-resource settings. ZACH-ViT contributes to this line of work by removing positional embeddings and the class token, reducing reliance on fixed spatial priors while maintaining an end-to-end patch-level transformer backbone. Its controlled MedMNIST evaluation established the regime-dependent validation used as the architectural reference in this work~\cite{angelakis2026zachvit}.

The original controlled ZACH-ViT benchmark compared compact scratch-trained models with larger pretrained CNN and Transformer families, including ABMIL, TransMIL, ResNet50, MobileNetV2, EfficientNetB0, DenseNet121, InceptionV3, DeiT-Small, Swin-Tiny, MambaOut-Tiny, and ConvNeXt-Tiny~\cite{ilse2018attention,shao2021transmil,he2016resnet,sandler2018mobilenetv2,tan2019efficientnet,huang2017densenet,szegedy2016inception,touvron2021deit,liu2021swin,zhang2024mambaout,liu2022convnext}. That benchmark also used Friedman/Nemenyi ranking analysis for multi-dataset comparisons~\cite{demsar2006statistical}. Related work outside medical imaging has also questioned whether positional embeddings should be treated as permanent architectural priors, for example in long-context language modeling~\cite{gelberg2025dropping}. hZACH-ViT does not re-open the entire backbone-comparison question; instead, it takes the verified compact ZACH-ViT backbone as fixed and studies the geometry of its final latent space.

\subsection{Permutation-Invariant Visual Representation Learning}

Permutation-invariant learning has a long history in set modeling and multiple-instance learning~\cite{ilse2018attention,shao2021transmil}. In medical imaging, multiple-instance learning is particularly common in histopathology, where whole-slide images are often represented as unordered bags of patches~\cite{shao2021transmil}. ZACH-ViT differs from classical multiple-instance learning aggregators because it is used as a compact image-level vision backbone rather than only as a bag-level aggregator over pre-extracted features. The present study extends this idea by asking whether the pooled representation should remain Euclidean or be embedded in a curved manifold. 

\subsection{Hyperbolic and Riemannian Representation Learning}

More generally, curved manifolds originate in the Riemannian view of geometry as a space with locally varying metric structure~\cite{riemann1868hypothesen}. Non-Euclidean representation learning has gained attention because many data domains exhibit hierarchical or curved structure~\cite{nickel2017poincare,ganea2018hyperbolic,chami2019hyperbolic,khrulkov2020hyperbolic}. Hyperbolic spaces, including the Poincar\'e ball and Klein model, have classical roots in non-Euclidean geometry~\cite{klein1871nichteuklidische,poincare1882groupes,cannon1997hyperbolic} and provide negatively curved geometries that can embed hierarchical relationships efficiently in modern representation learning~\cite{nickel2017poincare,ganea2018hyperbolic}. Spherical spaces, in contrast, provide positively curved geometries that may better organize angular or clustered relationships~\cite{liu2017sphereface,davidson2018hyperspherical}. While such geometries have been studied in natural language processing, graph learning, metric learning, and some computer vision settings, their role in compact medical Vision Transformers remains underexplored.

\subsection{Latent Geometry as an Architectural Design Axis}

Most medical imaging architectures focus on backbone design, pretraining, augmentation, or loss functions. Latent geometry is less commonly treated as an explicit architectural variable. This work studies latent geometry under a controlled setting: the backbone, data splits, optimization budget, and evaluation protocol remain fixed, while only the final representation geometry changes. This allows us to isolate whether non-Euclidean latent spaces provide benefits beyond standard Euclidean classification heads.

\section{Method}

\subsection{ZACH-ViT Backbone}

Let $X \in \mathbb{R}^{H \times W \times C}$ denote an input image. Following the original ZACH-ViT formulation~\cite{angelakis2026zachvit}, the backbone extracts non-overlapping patches of size $P \times P$ and flattens them into a sequence:
\begin{equation}
    Z_0 = \mathrm{ReLU}(\mathrm{Linear}(\mathrm{Patchify}(X))) \in \mathbb{R}^{N \times d},
\end{equation}
where $N=(H/P)(W/P)$ is the number of patches and $d$ is the embedding dimension. In the implementation used here, images are processed at $224\times224$ resolution with $P=16$, giving $N=196$ patches. Unlike standard ViTs~\cite{dosovitskiy2021image}, no positional embeddings are added and no class token is introduced~\cite{angelakis2026zachvit}. The patch sequence is processed by two transformer blocks with hidden dimensions $(128,64)$ and eight attention heads. The final image representation is obtained through global average pooling:
\begin{equation}
    h = \frac{1}{N}\sum_{i=1}^{N} Z_L^{(i)}.
\end{equation}
The pooled representation is passed through an MLP with hidden dimensions $(128,64)$ and dropout $0.1$. The resulting 64-dimensional vector is the input to either the Euclidean classifier or the non-Euclidean prototype head.

\subsection{Verified PyTorch Port}

The hZACH-ViT experiments use a PyTorch port of the original TensorFlow/Keras ZACH-ViT backbone. The notebook implements Keras-compatible Glorot initialization, LayerNorm with $\epsilon=10^{-6}$, and a Keras-like multi-head attention block to preserve the original architecture as closely as possible. The port includes a parameter-count audit. For the 8-class configuration, the expected TensorFlow ZACH-ViT count is 248,712 parameters and the PyTorch model contains exactly 248,712 parameters; for the binary configuration, both contain 248,257 parameters. This audit ensures that the Euclidean baseline and hZACH-ViT variants differ only in the final classifier head.

\subsection{hZACH-ViT Prototype Head}

hZACH-ViT preserves the ZACH-ViT backbone and modifies only the final latent geometry. Let $r \in \mathbb{R}^{64}$ denote the Euclidean representation produced by global average pooling and the MLP projection. The non-Euclidean head learns one Euclidean prototype vector per class, denoted $p_k \in \mathbb{R}^{64}$. Both the sample representation and class prototypes are mapped to a selected manifold:
\begin{equation}
    z = \Phi_{\mathcal{M},c}(r), \qquad q_k = \Phi_{\mathcal{M},c}(p_k),
\end{equation}
where $\mathcal{M}$ is the selected geometry and $c>0$ is the curvature magnitude. Throughout this work, $c$ denotes the magnitude of curvature: the hyperbolic heads have sectional curvature $-c$, while the spherical head has sectional curvature $+c$. Class logits are computed from manifold distances:
\begin{equation}
    \ell_k = -s\, d_{\mathcal{M},c}(z,q_k),
\end{equation}
where $s$ is a learnable non-negative scale parameter. For binary datasets, the model returns the logit difference $\ell_1-\ell_0$ and is trained with binary cross-entropy with logits. For multi-class datasets, all class logits are used with cross-entropy loss. The hZACH-ViT head adds $64K+1$ trainable parameters: $64\times K$ parameters for the class prototypes and one additional scale parameter. The backbone remains unchanged.

\subsection{Poincar\'e Ball Geometry}

The Poincar\'e ball model represents hyperbolic space as
\begin{equation}
    \mathbb{D}_{c}^{64} = \{z \in \mathbb{R}^{64}: c\|z\|^2 < 1\}.
\end{equation}
The implementation maps Euclidean vectors into the ball using
\begin{equation}
    \Phi_{\mathrm{Poincare},c}(u)=\tanh(\sqrt{c}\|u\|)\,\frac{u}{\sqrt{c}\|u\|},
\end{equation}
with the convention $\Phi_{\mathrm{Poincare},c}(0)=0$, followed by norm clipping to remain inside the ball of radius $1/\sqrt{c}$. The Poincar\'e distance between two mapped points $x$ and $y$ is
\begin{equation}
    d_{\mathbb{D}_c}(x,y)=\frac{1}{\sqrt{c}}\operatorname{arcosh}\!\left(1+\frac{2c\|x-y\|^2}{(1-c\|x\|^2)(1-c\|y\|^2)}\right).
\end{equation}

\subsection{Klein Geometry}

Poincar\'e and Klein coordinates represent the same underlying negatively curved geometry~\cite{cannon1997hyperbolic}, but the coordinate model and numerical behavior of the distance head differ. The Klein variant uses the same initial Poincar\'e mapping and then converts the coordinates to the Klein ball:
\begin{equation}
    x_K = \frac{2x_P}{1+c\|x_P\|^2}.
\end{equation}
For Klein coordinates $x$ and $y$, the distance is computed as
\begin{equation}
    d_{K_c}(x,y)=\frac{1}{\sqrt{c}}\operatorname{arcosh}\!\left(\frac{1-c\langle x,y\rangle}{\sqrt{(1-c\|x\|^2)(1-c\|y\|^2)}}\right).
\end{equation}
We therefore retain Klein as a coordinate-model control for the hyperbolic hypothesis, allowing us to test whether observed gains depend on negative curvature itself or on a particular Poincar\'e parameterization.

\subsection{Spherical Geometry}

The spherical variant provides a positive-curvature contrast. Euclidean vectors are projected to a sphere of radius $R=1/\sqrt{c}$.
\begin{equation}
    \Phi_{\mathrm{sphere},c}(u)=\frac{1}{\sqrt{c}}\frac{u}{\|u\|}.
\end{equation}
The spherical geodesic distance is
\begin{equation}
    d_{S_c}(x,y)=\frac{1}{\sqrt{c}}\arccos\!\left(\operatorname{clip}(c\langle x,y\rangle,-1,1)\right),
\end{equation}
where the clipping operation prevents floating-point round-off from moving the cosine argument outside $[-1,1]$. This head tests whether compact medical image representations benefit from positive rather than negative curvature.

\subsection{Curvature Sweep}

Curvature is not fixed a priori. The execution notebook evaluates
\begin{equation}
    c \in \{0.1, 0.2, 0.5, 1.0, 2.0, 5.0, 10.0\}
\end{equation}
for each of the three non-Euclidean geometries. The Euclidean ZACH-ViT baseline corresponds to flat geometry and is evaluated once per dataset and seed. With seven datasets, five seeds, one Euclidean baseline, three geometries, and seven curvatures, the planned sweep contains
\begin{equation}
    7 \times 5 \times (1 + 3\times 7)=770
\end{equation}
training runs. The notebook writes intermediate run summaries after each experiment, enabling interruption-safe execution and later aggregation.

\section{Experimental Setup}

\subsection{Datasets}

We evaluate seven datasets from MedMNIST v2~\cite{yang2023medmnist}: BloodMNIST, PathMNIST, BreastMNIST, PneumoniaMNIST, DermaMNIST, OCTMNIST, and OrganAMNIST. The datasets are loaded at $224\times224$ resolution using the official MedMNIST splits and RGB conversion. Pixel intensities are scaled to $[0,1]$. The same spatial-structure ordering as in the original ZACH-ViT study is retained: BloodMNIST is treated as the weakest spatial-structure regime, PathMNIST as weak/intermediate, BreastMNIST and PneumoniaMNIST as moderate, DermaMNIST as stronger but variable, and OCTMNIST and OrganAMNIST as the strongest anatomical-structure regimes~\cite{angelakis2026zachvit}.

\subsection{Few-Shot Protocol}

All experiments follow a controlled few-shot protocol implemented in the uploaded execution notebook:
\begin{itemize}[leftmargin=*]
    \item 50 training samples per class sampled from the official training split without replacement;
    \item official validation and test splits kept unchanged;
    \item five random seeds: $\{3,5,7,11,13\}$;
    \item image size: $224 \times 224$ with RGB input;
    \item patch size: $16\times16$;
    \item batch size: 16;
    \item optimizer: Adam with learning rate $10^{-4}$, $\epsilon=10^{-8}$, and no weight decay;
    \item training epochs for the full geometry--curvature sweep: 47;
    \item number of workers: 0, to maximize deterministic behavior in the notebook environment.
\end{itemize}

\begin{table}[t]
\centering
\caption{Notebook-derived model and experiment configuration used for the hZACH-ViT sweep.}
\label{tab:notebook_config}
\begin{tabular}{ll}
\toprule
Component & Configuration \\
\midrule
Backbone & ZACH-ViT-Torch, verified PyTorch port \\
Input size & $224\times224\times3$ \\
Patch size & $16\times16$ \\
Patch tokens & 196 \\
Transformer units & $(128,64)$ \\
Attention heads & 8 \\
MLP head units & $(128,64)$ \\
Dropout & 0.1 \\
Latent dimension for geometry head & 64 \\
Euclidean classifier & Linear layer \\
hZACH classifier & Class prototypes + learnable distance scale \\
Geometries & Poincar\'e, Klein, spherical \\
Curvatures & $0.1,0.2,0.5,1.0,2.0,5.0,10.0$ \\
Seeds & $3,5,7,11,13$ \\
Epochs & 47 \\
Planned runs & 770 \\
\bottomrule
\end{tabular}
\end{table}

\subsection{Models Compared}

The main comparison includes:
\begin{itemize}[leftmargin=*]
    \item \textbf{ZACH-ViT-Torch}: verified Euclidean PyTorch implementation of ZACH-ViT;
    \item \textbf{hZACH-ViT-Poincar\'e}: negatively curved Poincar\'e prototype head;
    \item \textbf{hZACH-ViT-Klein}: negatively curved Klein prototype head;
    \item \textbf{hZACH-ViT-Spherical}: positively curved spherical prototype head.
\end{itemize}
The original TensorFlow ZACH-ViT is used as architectural reference, while the final comparisons use the verified PyTorch baseline so that all geometry variants are evaluated under the same software framework.

\subsection{Metrics}

For multi-class datasets, the primary metric is test MacroF1. For binary datasets, the notebook reports probability-based AUC, thresholded AUC@0.5 based on predicted labels, F1, accuracy, sensitivity, and specificity. To maintain continuity with the original ZACH-ViT benchmark, the primary binary score used by the notebook's aggregation cell is thresholded AUC@0.5, while probability AUC is retained as a supplementary metric~\cite{angelakis2026zachvit}. Results are aggregated as mean $\pm$ standard deviation over five seeds.

\section{Results}

\subsection{Full Sweep Completion}

The latest full results file contains the complete planned sweep: 770 runs, corresponding to seven datasets, five random seeds, one Euclidean ZACH-ViT baseline, and 21 non-Euclidean hZACH-ViT settings per seed. Each dataset contains 110 runs: $5$ seeds $\times$ $(1 + 3 \times 7)$ geometry--curvature settings. OCTMNIST and OrganAMNIST are both included, and no dataset is left empty.

\subsection{Euclidean ZACH-ViT versus Best hZACH-ViT}

Table~\ref{tab:completed_results} reports the completed seven-dataset comparison between the verified Euclidean ZACH-ViT PyTorch baseline and the best hZACH-ViT setting for each non-Euclidean geometry. For multi-class datasets, the score is test MacroF1. For binary datasets, the score is thresholded test AUC@0.5, following the aggregation convention used in the execution notebook and the original ZACH-ViT benchmark. All values are mean $\pm$ standard deviation over five seeds.

\begin{table}[t]
\centering
\scriptsize
\caption{Completed comparison between Euclidean ZACH-ViT and hZACH-ViT geometry heads across MedMNIST7. The best hZACH-ViT setting is selected within each dataset from the full 21-setting non-Euclidean sweep (three geometries $\times$ seven curvatures); therefore, this table reports a model-selection result rather than a fixed-head comparison.}
\label{tab:completed_results}
\resizebox{\textwidth}{!}{%
\begin{tabular}{llllllll}
\toprule
Dataset & Metric & Euclidean ZACH-ViT & Best Poincar\'e & Best Klein & Best spherical & Best hZACH-ViT & $\Delta$ vs. Euclidean \\
\midrule
BloodMNIST & MacroF1 & 0.790 $\pm$ 0.035 & 0.786 $\pm$ 0.041 ($c=0.1$) & 0.778 $\pm$ 0.044 ($c=0.1$) & 0.793 $\pm$ 0.012 ($c=0.1$) & spherical, $c=0.1$ & +0.003 \\
PathMNIST & MacroF1 & 0.577 $\pm$ 0.045 & 0.585 $\pm$ 0.081 ($c=0.5$) & 0.588 $\pm$ 0.062 ($c=0.1$) & 0.603 $\pm$ 0.060 ($c=0.2$) & spherical, $c=0.2$ & +0.026 \\
BreastMNIST & AUC@0.5 & 0.624 $\pm$ 0.059 & 0.643 $\pm$ 0.042 ($c=0.2$) & 0.632 $\pm$ 0.040 ($c=0.1$) & 0.639 $\pm$ 0.048 ($c=0.2$) & Poincar\'e, $c=0.2$ & +0.019 \\
PneumoniaMNIST & AUC@0.5 & 0.719 $\pm$ 0.016 & 0.725 $\pm$ 0.019 ($c=0.1$) & 0.728 $\pm$ 0.034 ($c=1.0$) & 0.734 $\pm$ 0.020 ($c=5.0$) & spherical, $c=5.0$ & +0.015 \\
DermaMNIST & MacroF1 & 0.309 $\pm$ 0.023 & 0.325 $\pm$ 0.021 ($c=0.2$) & 0.312 $\pm$ 0.019 ($c=0.1$) & 0.328 $\pm$ 0.034 ($c=0.1$) & spherical, $c=0.1$ & +0.019 \\
OCTMNIST & MacroF1 & 0.274 $\pm$ 0.047 & 0.329 $\pm$ 0.055 ($c=0.1$) & 0.328 $\pm$ 0.028 ($c=5.0$) & 0.326 $\pm$ 0.041 ($c=1.0$) & Poincar\'e, $c=0.1$ & +0.055 \\
OrganAMNIST & MacroF1 & 0.499 $\pm$ 0.033 & 0.511 $\pm$ 0.025 ($c=0.2$) & 0.498 $\pm$ 0.019 ($c=0.1$) & 0.488 $\pm$ 0.041 ($c=0.1$) & Poincar\'e, $c=0.2$ & +0.012 \\
\bottomrule
\end{tabular}}

\end{table}

\begin{figure}[t]
\centering
\includegraphics[width=0.98\textwidth]{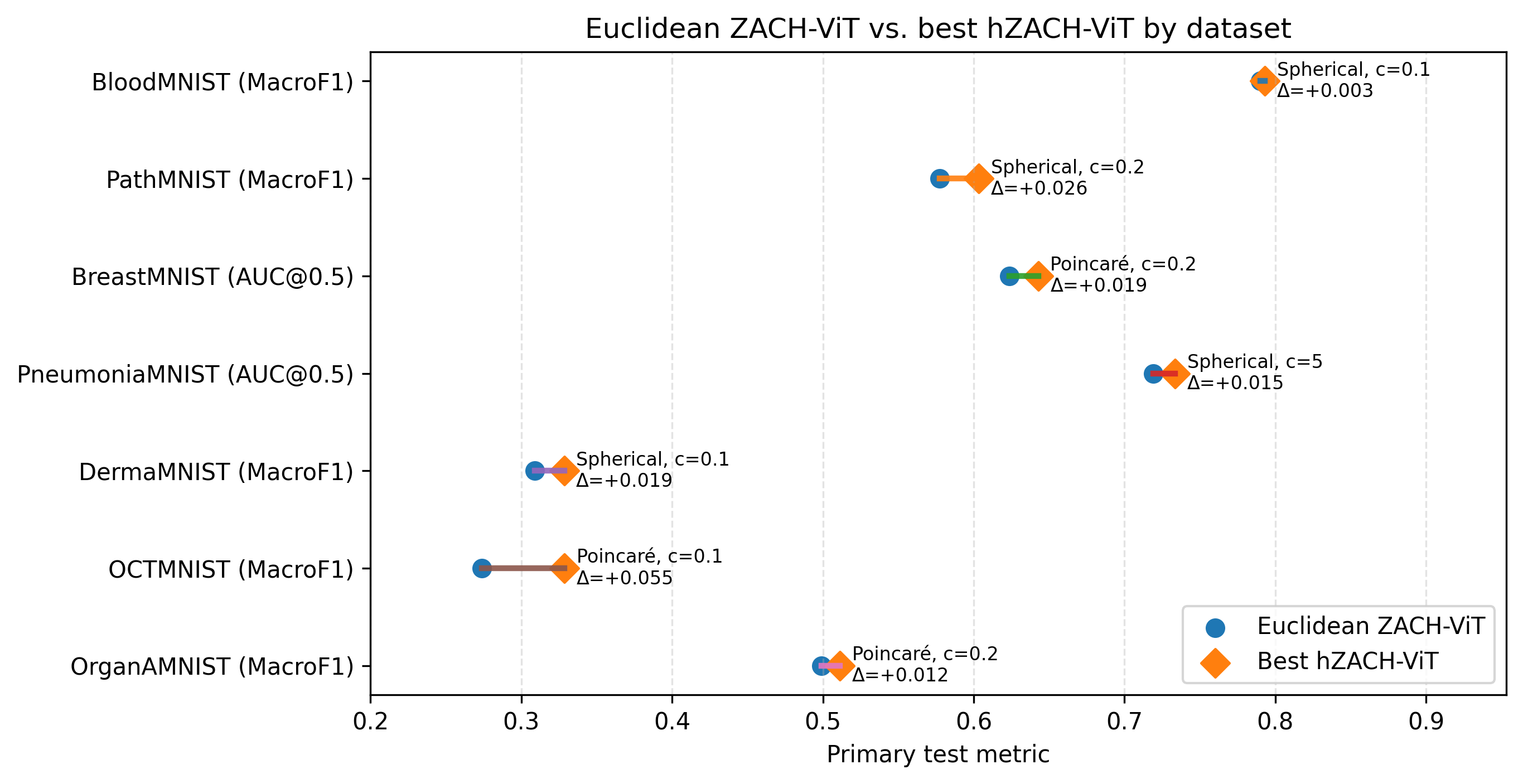}
\caption{Dataset-wise comparison between the verified Euclidean ZACH-ViT baseline and the best hZACH-ViT configuration selected from the full geometry--curvature sweep. Each line links the Euclidean baseline to the best non-Euclidean result for the same dataset, and the annotation reports the selected geometry, curvature, and absolute gain.}
\label{fig:euclid_vs_best}
\end{figure}

Across all seven datasets, the best non-Euclidean hZACH-ViT head improves over the Euclidean ZACH-ViT baseline in the completed sweep.
The mean absolute gain across datasets is $+0.021$ in the corresponding primary metric, with improvements ranging from $+0.003$ on BloodMNIST to $+0.055$ on OCTMNIST. The gains are not uniform, but the direction is consistent across the seven-dataset benchmark.

The strongest gain is observed on OCTMNIST, where Poincar\'e geometry at $c=0.1$ improves the test MacroF1 from $0.274\pm0.047$ to $0.329\pm0.055$. PathMNIST also shows a clear improvement, with the spherical head at $c=0.2$ increasing MacroF1 from $0.577\pm0.045$ to $0.603\pm0.060$. The smallest gain appears on BloodMNIST, where spherical geometry at $c=0.1$ yields $0.793\pm0.012$ compared with $0.790\pm0.035$ for Euclidean ZACH-ViT.

\subsection{Statistical and Model-Selection Analysis}

Because each dataset-level hZACH-ViT result is selected post-hoc from 21 non-Euclidean configurations (three geometries and seven curvatures), Table~\ref{tab:completed_results} should be interpreted as a model-selection upper bound rather than as a single fixed-head comparison. This point is important for fair interpretation: the consistency of the direction of improvement is encouraging, but the selected configuration benefits from a larger search space than the Euclidean baseline.

To quantify seed-level uncertainty, Table~\ref{tab:stats_tests} reports paired comparisons between Euclidean ZACH-ViT and the selected hZACH-ViT configuration for each dataset. With only five seeds, individual paired tests have low power. Consequently, only PathMNIST reaches $p < 0.05$ under a paired $t$-test, while OCTMNIST shows the largest absolute gain but remains above the conventional threshold. BloodMNIST is appropriately interpreted as a negligible gain. Several effects are moderate to large in magnitude despite non-significant p-values under $n=5$ seeds; for example, OCTMNIST has the largest gain ($+0.055$) and a large paired effect size ($d_z=0.86$), but the paired $t$-test remains underpowered. Across datasets, however, all seven differences are positive; a two-sided sign test for 7/7 positive effects gives $p = 0.0156$. Given the small number of seeds, larger-scale replication with 10--20 seeds would be needed to detect small-effect improvements reliably, but the consistent direction of effects and the fixed-curvature analysis support the practical relevance of the trend.

\begin{table}[t]
\centering
\scriptsize
\caption{Seed-level exploratory comparison between Euclidean ZACH-ViT and the selected hZACH-ViT configuration. $\Delta$ is the paired mean difference over five seeds; the 95\% confidence interval is a bootstrap interval over seeds. Because the selected hZACH-ViT head is chosen from the full geometry--curvature sweep and each comparison uses only five seeds, these p-values should be interpreted descriptively rather than confirmatorily.}
\label{tab:stats_tests}
\resizebox{\textwidth}{!}{%
\begin{tabular}{llllll}
\toprule
Dataset & Selected hZACH-ViT & $\Delta$ [95\% CI] & Paired $t$ $p$ & Wilcoxon $p$ & Cohen's $d_z$ \\
\midrule
BloodMNIST & spherical, $c=0.1$ & +0.003 [$-0.018$, +0.028] & 0.817 & 1.000 & 0.11 \\
PathMNIST & spherical, $c=0.2$ & +0.026 [+0.012, +0.035] & 0.019 & 0.125 & 1.71 \\
BreastMNIST & Poincar\'e, $c=0.2$ & +0.019 [$-0.032$, +0.070] & 0.553 & 0.625 & 0.29 \\
PneumoniaMNIST & spherical, $c=5.0$ & +0.015 [+0.001, +0.027] & 0.122 & 0.313 & 0.88 \\
DermaMNIST & spherical, $c=0.1$ & +0.019 [$-0.014$, +0.059] & 0.399 & 0.313 & 0.42 \\
OCTMNIST & Poincar\'e, $c=0.1$ & +0.055 [+0.015, +0.112] & 0.127 & 0.063 & 0.86 \\
OrganAMNIST & Poincar\'e, $c=0.2$ & +0.012 [$-0.018$, +0.037] & 0.472 & 0.625 & 0.35 \\
\bottomrule
\end{tabular}}
\end{table}

We also performed a dataset-level Friedman test comparing four families: Euclidean ZACH-ViT, best Poincar\'e over curvature, best Klein over curvature, and best spherical over curvature. The mean ranks were spherical 1.86, Poincar\'e 2.00, Klein 2.71, and Euclidean 3.43, but the Friedman test did not reach significance ($\chi^2=6.60$, $p = 0.086$). This supports a cautious interpretation: the sweep reveals a consistent and practically relevant trend, but not enough evidence to claim statistically significant superiority of one geometry family across seven datasets.

\subsection{Fixed-Curvature Robustness Analysis}

To address the multiple-comparisons concern, we performed an additional fixed-curvature analysis using the already completed sweep, without training new models. Table~\ref{tab:fixed_curvature} summarizes three increasingly restrictive protocols. First, the full per-dataset sweep allows selection over all 21 non-Euclidean settings. Second, a fixed-curvature protocol fixes $c=0.1$ globally and selects only the geometry per dataset. Third, a single global configuration uses the same geometry and curvature for every dataset. The fixed $c=0.1$ protocol remains positive on all seven datasets with a mean gain of $+0.016$, and the single global spherical $c=0.1$ setting remains positive on five of seven datasets with a mean gain of $+0.011$. Thus, the main conclusion is not solely dependent on exhaustive 21-way selection.

\begin{table}[t]
\centering
\caption{Robustness to reduced model selection. The primary full-sweep result is compared with fixed-curvature and single-global-configuration protocols.}
\label{tab:fixed_curvature}
\begin{tabular}{llll}
\toprule
Protocol & Selection allowed & Mean $\Delta$ & Positive datasets \\
\midrule
Full hZACH-ViT sweep & geometry + curvature per dataset & +0.021 & 7 / 7 \\
Fixed $c=0.1$ & geometry per dataset only & +0.016 & 7 / 7 \\
Fixed $c=0.2$ & geometry per dataset only & +0.014 & 6 / 7 \\
Single global spherical $c=0.1$ & none after choosing global setting & +0.011 & 5 / 7 \\
Single global Poincar\'e $c=0.1$ & none after choosing global setting & +0.010 & 5 / 7 \\
\bottomrule
\end{tabular}
\end{table}

\subsection{Geometry-Specific Best Curvatures}

Spherical geometry is the best overall non-Euclidean head for four datasets (BloodMNIST, PathMNIST, PneumoniaMNIST, and DermaMNIST), while Poincar\'e geometry is best for BreastMNIST, OCTMNIST, and OrganAMNIST. Klein geometry is competitive in several settings, particularly OCTMNIST, but it is not the top geometry in the completed sweep.

The completed results support the central premise of hZACH-ViT: changing only the final latent geometry and prototype-distance head can systematically affect performance while preserving the compact ZACH-ViT backbone.
The results also indicate that the optimal geometry is dataset dependent. Positively curved spherical geometry performs best in several weak-to-moderate structure regimes, whereas Poincar\'e geometry is strongest for BreastMNIST, OCTMNIST, and OrganAMNIST. This suggests that latent geometric bias is a tunable design axis rather than a universally fixed choice.

\subsection{Interpretation of the Completed Sweep}

The completed 770-run sweep should be interpreted as a controlled curvature-tuning study rather than as evidence that one manifold universally dominates. The best hZACH-ViT variant improves over Euclidean ZACH-ViT on every dataset, but this comparison allows dataset-specific selection over three geometries and seven curvatures. Consequently, the main conclusion is not that a single non-Euclidean geometry should replace Euclidean heads in all compact ViTs. Rather, the result shows that non-Euclidean prototype heads provide a reproducible and empirically useful extension of ZACH-ViT, and that the curvature/geometry choice should be treated as a model-selection variable in low-data medical imaging.

\begin{figure}[t]
\centering
\includegraphics[width=0.98\textwidth]{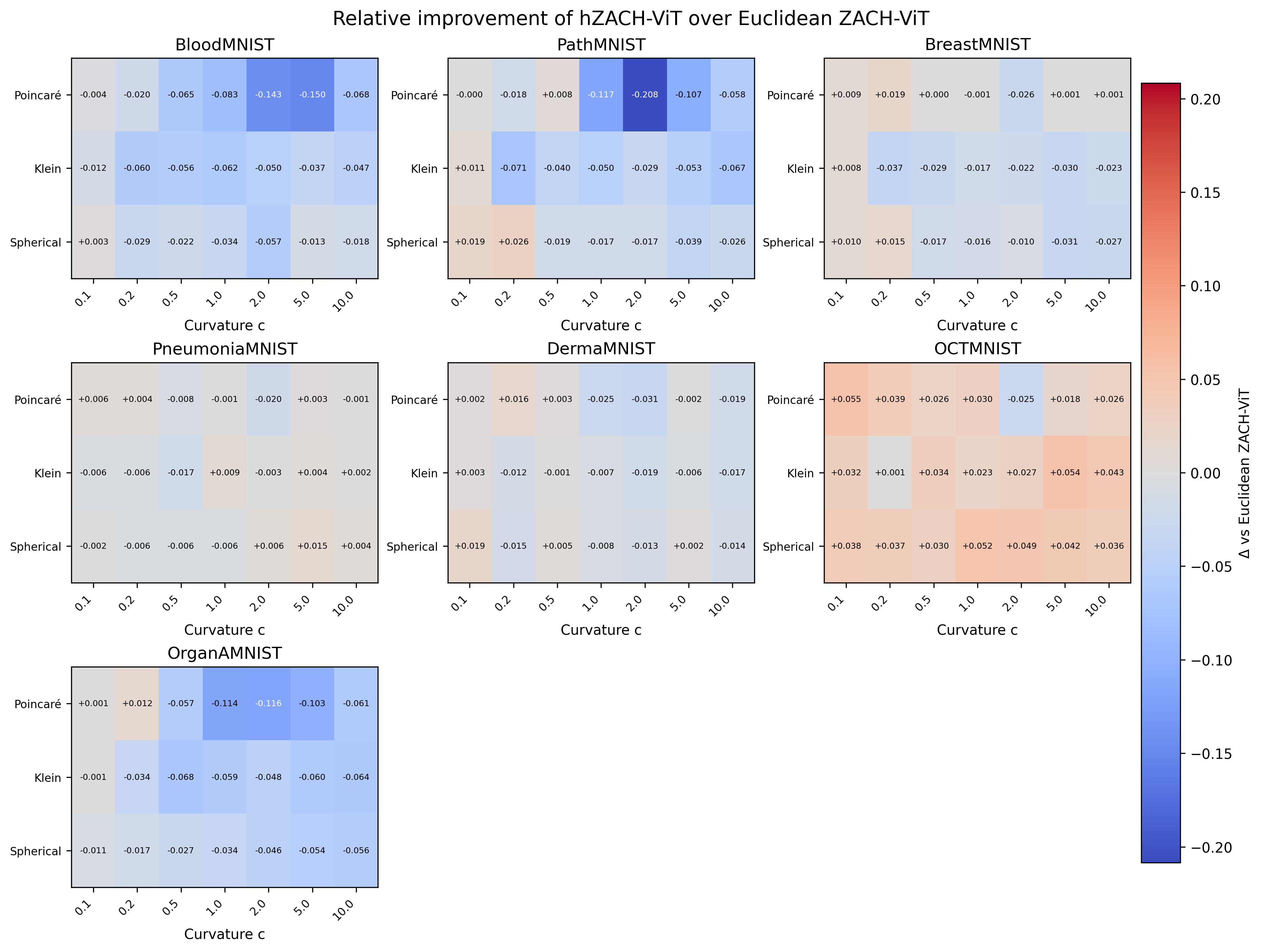}
\caption{Relative improvement of hZACH-ViT over Euclidean ZACH-ViT for every geometry--curvature setting. Positive values indicate gains over the Euclidean baseline and negative values indicate losses, using the dataset-specific primary metric. This visualization makes the model-selection landscape explicit and shows that useful improvements are not confined to a single manifold family.}
\label{fig:delta_heatmaps}
\end{figure}

\section{Discussion}

The completed seven-dataset results support the hypothesis that latent geometry is a meaningful design variable for compact medical Vision Transformers. Figure~\ref{fig:euclid_vs_best} shows the absolute dataset-level gains, while Figure~\ref{fig:delta_heatmaps} shows the full geometry--curvature response surface. Importantly, the evidence does not imply that one non-Euclidean geometry is universally optimal. Instead, the observed pattern is more nuanced: compact permutation-invariant representations can benefit from curved latent spaces, but the optimal curvature and geometry depend on the imaging domain. In practical terms, every dataset benefits in the completed sweep, but the magnitude of improvement and the winning manifold family vary materially across tasks.

This observation is consistent with the regime-dependent behavior of the original ZACH-ViT architecture~\cite{angelakis2026zachvit}. ZACH-ViT showed that the usefulness of positional priors depends on the spatial structure of the image domain. hZACH-ViT extends this principle from spatial inductive bias to latent geometric bias. Just as positional priors are not universally beneficial, Euclidean latent geometry may not be universally optimal.

The completed sweep suggests two practical conclusions. First, spherical geometry is a strong candidate default among the tested non-Euclidean heads, achieving the best hZACH-ViT result on four of seven datasets. Second, Poincar\'e geometry remains important, achieving the best hZACH-ViT result on BreastMNIST, OCTMNIST, and OrganAMNIST. This split is scientifically useful because it argues against a simplistic ``hyperbolic is always better'' interpretation. Instead, the manifold should be selected or validated as part of the model design process.

A speculative interpretation is that spherical geometry may be well matched to datasets whose classes form compact, angularly separated clusters in the ZACH-ViT representation space, such as blood cell categories, histopathology patterns, and lesion-like or infection-related appearances. Poincar\'e geometry may instead be more helpful when class relationships contain hierarchical or nested structure, such as organ categories or retinal-layer disease patterns. The present experiments do not prove this mechanism, but they motivate future latent-space visualization and class-hierarchy analysis.

The low-curvature pattern is also important. Six of the seven dataset-level winners occur at $c=0.1$ or $c=0.2$. Mild curvature may act as a soft geometric regularizer: it reshapes prototype decision boundaries while preserving much of the useful local Euclidean neighborhood structure learned by the backbone. Larger curvature values may distort the 64-dimensional representation too aggressively, which is consistent with the degradation observed in several high-curvature settings.

Although Klein geometry never wins a dataset in the final selection, we retain it in the main analysis because it provides a useful coordinate-model control for the hyperbolic hypothesis. Poincar\'e and Klein coordinates represent the same underlying negatively curved geometry but lead to different numerical distance computations. The fact that Klein is competitive on OCTMNIST but not generally dominant suggests that the observed gains are not simply due to adding any hyperbolic coordinate transform; the coordinate model and numerical behavior matter. Klein's competitiveness on OCTMNIST also suggests that stable manifold parameterizations deserve future study.

The OCTMNIST result is particularly noteworthy. OCTMNIST is one of the strongest spatial-structure regimes in the benchmark, and the completed sweep shows the largest gain there. The relative-improvement heatmap in Figure~\ref{fig:delta_heatmaps} further shows that OCTMNIST benefits from a relatively broad band of strong non-Euclidean settings, not just a single isolated point. This indicates that the benefit of hZACH-ViT is not restricted to weakly ordered microscopy-like data. A plausible interpretation is that the non-Euclidean prototype head improves class separation in the final representation space even when the visual input has stable anatomical organization. This point should be examined further with latent-space visualization and calibration analyses.

\subsection{Practical Takeaways}

For practitioners, the results suggest a simple use pattern. If a full curvature sweep is affordable, hZACH-ViT can be treated as a lightweight model-selection layer on top of a fixed compact backbone. If only one or two settings can be tested, spherical geometry with $c=0.1$ is a reasonable first default, followed by Poincar\'e geometry at $c=0.1$ or $c=0.2$ when the task appears hierarchical or anatomically structured. The overhead is minimal in parameter count: for $K$ classes, the prototype head uses $64K+1$ trainable parameters, which is the same order as the Euclidean linear classifier and negligible relative to the approximately 0.25M-parameter ZACH-ViT backbone. The additional computation is limited to mapping the final 64-dimensional representation and class prototypes to the selected manifold and computing prototype distances. In the present notebook timing logs, the non-Euclidean heads increased end-to-end test-time wall-clock inference by approximately 19\% on average. This estimate comes from the research notebook timing path rather than from a dedicated GPU-optimized or JIT-compiled deployment kernel; optimized 64-dimensional manifold-distance kernels would likely reduce this overhead.

\section{Limitations}

The most important limitation is model selection and multiple comparisons: the main hZACH-ViT result selects the best configuration from 21 non-Euclidean settings per dataset. The fixed-curvature analysis mitigates this concern by showing that low-curvature configurations retain positive gains under reduced selection, but it does not replace nested validation or an external hold-out protocol. Future versions should therefore select geometry and curvature on validation data before final testing.

A second limitation is that the Euclidean baseline uses a standard linear classifier rather than a Euclidean prototype head with learnable scale. Therefore, the reported gains reflect the combination of curved geometry and prototype-based distance classification, not geometry alone. The fixed-curvature robustness analysis suggests that geometry contributes materially to the improvement, but a future Euclidean-prototype control would isolate this effect more cleanly.

A third limitation is scope. The study is restricted to MedMNIST-style benchmark datasets and a controlled few-shot protocol. Although this setting is valuable for isolating architectural and geometric effects, it does not fully represent the complexity of raw clinical imaging, raw DICOM workflows, or 3D volumes. Future work should validate hZACH-ViT on larger clinical datasets, raw DICOM pipelines, volumetric imaging, and external multi-center cohorts.

Finally, the current analysis focuses on classification performance and seed-level stability. Future work should include calibration analysis, latent-space visualization, decision-boundary diagnostics, and larger replication studies with more random seeds to better characterize when and why curved prototype heads improve compact Vision Transformers.

\section{Conclusion}

We introduced hZACH-ViT, a family of curved-latent-space extensions of the compact ZACH-ViT architecture. By preserving the verified ZACH-ViT backbone and varying only the final geometry-aware prototype head, we provide a controlled study of Euclidean, hyperbolic, and spherical representations in low-data medical imaging. The completed 770-run MedMNIST7 geometry--curvature sweep shows that the best non-Euclidean hZACH-ViT head improves over Euclidean ZACH-ViT on all seven datasets, with the largest gain observed on OCTMNIST and an average gain of $+0.021$ across the benchmark. The preferred manifold is dataset dependent: spherical geometry is strongest on four datasets, while Poincar\'e geometry is strongest on three, and low curvature dominates most winners. Overall, these findings position latent geometry as a controlled, reproducible, and practically useful design axis for compact medical Vision Transformers, and make hZACH-ViT a natural fit for computer-vision settings where geometry is treated as an explicit part of model design rather than an implicit default.

\bibliographystyle{unsrt}
\bibliography{references}
\end{document}